\documentclass{article}

\usepackage{microtype}
\usepackage{graphicx}
\usepackage{subcaption}
\usepackage{booktabs} 

\usepackage{makecell}
\usepackage{subcaption}
\usepackage{caption}
\usepackage{framed}  
\usepackage[font=small,skip=0pt]{caption}
\usepackage[flushleft]{threeparttable}

\usepackage{pdfpages}
\usepackage{hyperref}

\usepackage{times}
\usepackage{epsfig}
\usepackage{graphicx}
\usepackage{amsmath}
\usepackage{amssymb}
\usepackage{bbm}
\usepackage{dsfont}

\usepackage{MnSymbol}

\usepackage[accepted]{icml2020}

\icmltitlerunning{XtarNet: Learning to Extract Task-Adaptive Representation for Incremental Few-Shot Learning}

\begin{document}

\twocolumn[
\icmltitle{XtarNet: Learning to Extract Task-Adaptive Representation \\ for Incremental Few-Shot Learning}




\begin{icmlauthorlist}
\icmlauthor{Sung Whan Yoon$^{*}$}{unist}
\icmlauthor{Do-Yeon Kim$^{*}$}{kaist}
\icmlauthor{Jun Seo}{kaist}
\icmlauthor{Jaekyun Moon}{kaist}

\end{icmlauthorlist}

\icmlaffiliation{kaist}{School of Electrical Engineering, Korea Advanced Institute of Science and Technology (KAIST), Daejeon, South Korea}
\icmlaffiliation{unist}{School of Electrical and Computer Engineering, Ulsan National Institute of Science and Technology (UNIST), Ulsan, South Korea}

\icmlcorrespondingauthor{Sung Whan Yoon}{shyoon8@unist.ac.kr}

\icmlkeywords{Machine Learning, ICML}

\vskip 0.3in
]



\printAffiliationsAndNotice{\icmlEqualContribution}  

\begin{abstract}

Learning novel concepts while preserving prior knowledge is a long-standing challenge in machine learning. The challenge gets greater when a novel task is given with only a few labeled examples, a problem known as \textit{incremental few-shot learning}. We propose \textit{XtarNet}, which learns to extract task-adaptive representation (TAR) for facilitating incremental few-shot learning. The method utilizes a backbone network pretrained on a set of base categories while also employing additional modules that are meta-trained across episodes. Given a new task, the novel feature extracted from the meta-trained modules is mixed with the base feature obtained from the pretrained model. The process of combining two different features provides TAR and is also controlled by meta-trained modules. The TAR contains effective information for classifying both novel and base categories. The base and novel classifiers quickly adapt to a given task by utilizing the TAR. Experiments on standard image datasets indicate that XtarNet achieves state-of-the-art incremental few-shot learning performance. The concept of TAR can also be used in conjunction with existing incremental few-shot learning methods; extensive simulation results in fact show that applying TAR enhances the known methods significantly.

\end{abstract}

\section{Introduction}
Humans can quickly learn novel concepts from limited experience.
In contrast, deep learning usually requires a massive amount of training data to learn each novel task. 
Unfortunately, gathering sufficient training samples in many applications can be highly expensive and impractical. 
To tackle this issue in computer vision, many researchers have dedicated to develop few-shot learning algorithms, which aim to learn generalized models for classifying unseen categories with only a few labeled images.
A popular way to develop few-shot learners is to adopt meta-training in \textit{episodic} form \cite{MN}, where the learners are exposed to widely-varying episodes one by one, each time with a few-labeled images.
Built upon the episodic training, successful few-shot learning algorithms attempt to build inductive bias over varying tasks while employing task-dependent adaptation for each task \cite{PN, MAML, SNAIL, TADAM, TapNet}.

\begin{figure*}
	\centering
	\begin{subfigure}[b]{0.55\textwidth}
		\includegraphics[width=\textwidth]{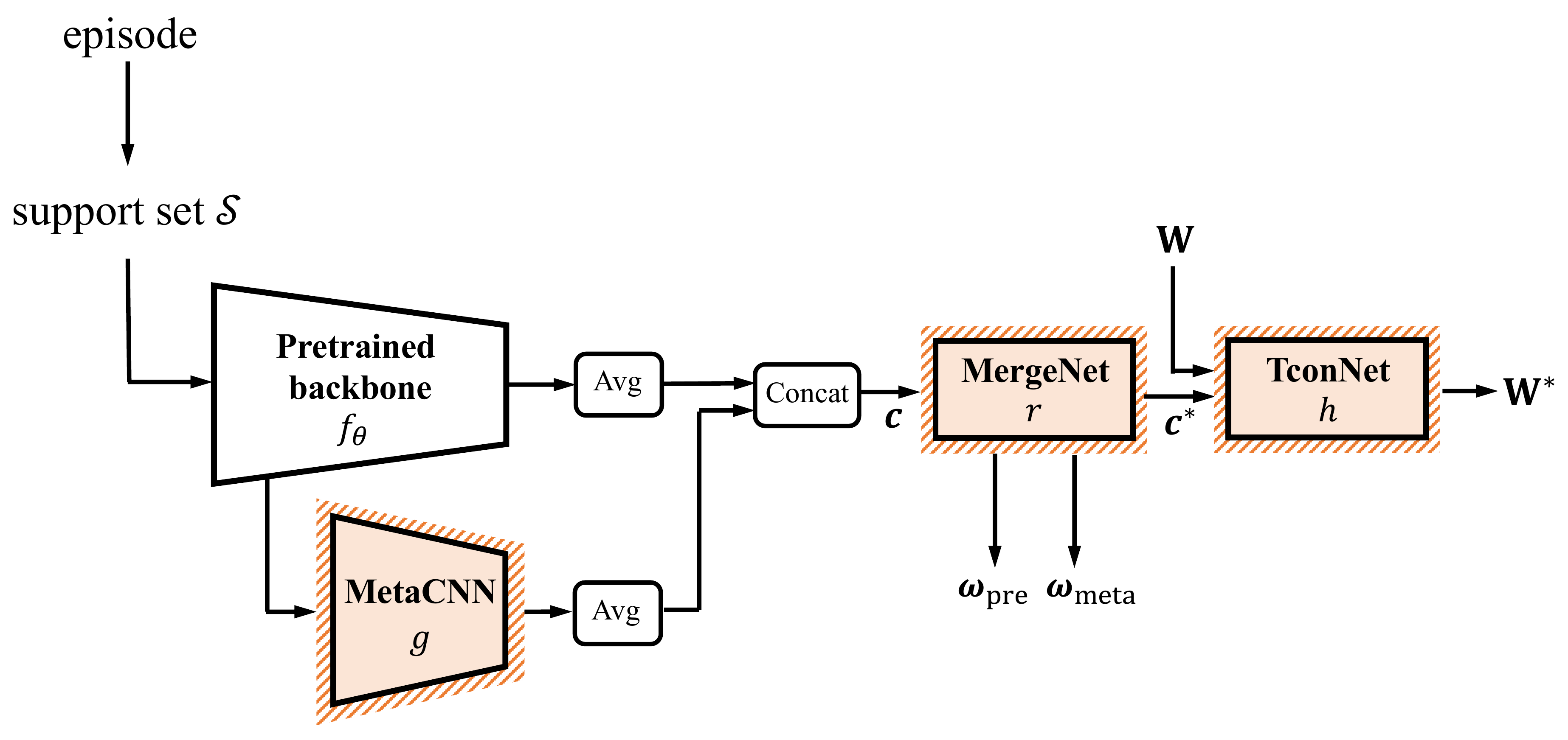}
		\caption{Processing of a support set}
		\label{fig:diagram1}
	\end{subfigure}	
	\begin{subfigure}[b]{0.43\textwidth}
		\includegraphics[width=\textwidth]{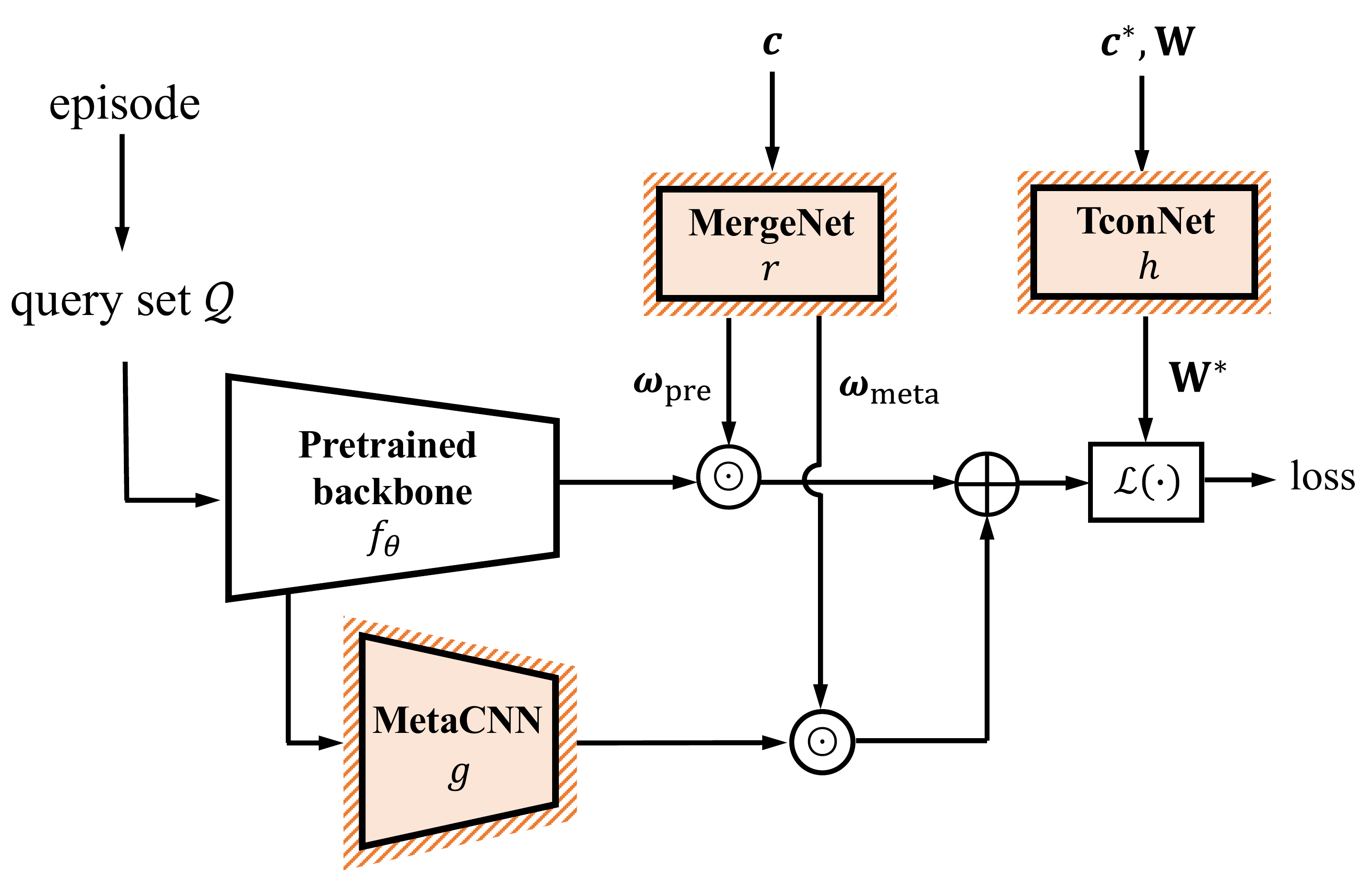}
		\caption{Processing of a query set}
		\label{fig:diagram2}
	\end{subfigure}	
	\caption{XtarNet processing of a given episode: our modules can be plugged in with existing incremental few-shot learners, e.g., Imprint and LwoF. Although TapNet was not proposed as an incremental few-shot learner, it can be used as an effective novel classifier for our purposes, in which case projection M is computed such that per-class average of the combined features and novel classifier weights coincide in M.}
	\label{fig:diagram}
\end{figure*}

While well-known few-shot learning methods focus mainly on classifying novel categories, a recent line of works attempts to handle novel categories while retaining the ability to classify a set of original base categories \cite{dynamic, incMeta}. 
This more recent problem is referred to as incremental few-shot learning.
Here, based on a prepared model which is trained on a set of original base categories, extra novel categories
should be quickly learned within a few shots.
To this end, incremental few-shot learners strive to build a generalized model while preventing catastrophic forgetting of the base categories. 
Achieving this objective is extremely arduous; simple extensions of existing few-shot learning algorithms to the setting
of incremental few-shot learning does not work.
Built upon a pretrained backbone, the recent methods of \cite{dynamic, incMeta} focus on learning to learn a set of new classification weights for novel categories to balance classification performance on base and novel categories. 
The attention-based method of \cite{dynamic} learns to leverage the past knowledge by generating novel class weights with a form of attention over the pretrained classification weights for base categories. 
A more recent method of \cite{incMeta} called Attention Attractor Network employs an inner loop of optimization for generating a novel classifier for each episode. To be specific, meta-learned modules compute an attractor vector, and then a regularization term based on the distance between the attractor and novel classification weights is used in the inner loop optimization.
The prior methods which learn to build a novel classifier show substantial performance gains over the baseline approach that uses per-class averages of embedded support samples as novel classification weights.
However, these algorithms focus on extracting a \textit{fixed} representation from the pretrained backbone, rather than exploring novel representations that could be task-adaptive.

Our motivation here is to pursue novel knowledge from new experience and merge it with prior knowledge learned from previous experience.
In this paper, we propose \textit{XtarNet}, which learns to construct novel representation with informative features to classify both base and novel categories.
In addition to a pretrained feature extractor, our method employs extra modules that are meta-trained across varying episodes to accomplish the following three objectives: design of a novel feature extractor that can be trained easily across different tasks, construction of \textit{task-adaptive representation} (TAR) by mixing the novel and base features together, and provision of efficient task-conditioning of base and novel classifiers by utilizing the TAR.

Experiments on incremental few-shot classification tasks with \textit{mini}ImageNet and \textit{tiered}ImageNet show that XtarNet
achieves the state-of-the-art performance. The proposed TAR idea can be used in conjunction with different incremental few-shot learning methods, improving classification accuracies of existing learners.

\section{Proposed Model}

We follow the incremental few-shot learning setting proposed in \cite{dynamic, incMeta}. First, a backbone network is pretrained on a set of base categories. Built upon the pretrained model, meta-training is done with episodic form of training.
For each episode of incremental few-shot learning, a few labeled samples (i.e., support samples) from novel categories are given to the model. Based on the support set, the model aims to classify query samples from both base and novel categories.

Our XtarNet utilizes three different meta-learnable modules in addition to a backbone network. Fig. \ref{fig:diagram1} describes how XtarNet processes the support set $\mathcal{S}$ of a given episode. The backbone feature extractor $f_{\theta}$ is prepared through a regular supervised learning on a training dataset for base categories. The classification weight vector set $\mathbf{W}$ consists of base classifier weight vectors, which are also pretrained on base categories, as well as novel classifier weight vectors, which vary with each training episode.
\textit{MetaCNN} $g$ is a small convolutional neural netwok (CNN) module that takes as input an intermediate layer output of $f_{\theta}$. The output of $g$ represents the \textit{novel feature}. Per-class average features are collected at the outputs of both backbone and MetaCNN. The backbone network output represents the \textit{base feature}. 
\textit{MergeNet} $r$, consisting of two relatively small fully-connected networks, creates a mixture of the base and novel features.
This mixture is what we call the task-adaptive representation (TAR) of the given input image. 
\textit{TconNet} $h$, based on another set of small fully-connected networks, provides task-conditioning on the classification weights $\mathbf{W}$ by utilizing the TAR.   

Fig. \ref{fig:diagram2} depicts processing of a query set and the learning process. Once the support set is processed, the mixture weight vectors, 
$\boldsymbol{\omega}_{\text{pre}}$ and $\boldsymbol{\omega}_{\text{meta}}$, as well as the conditioned weights $\mathbf{W}^{*}$ become available to be used during the query processing. The given query image is processed again through both backbone and MetaCNN, and the output features are combined via the given weights as shown. Cross entropy loss is computed using the classifier $\mathbf{W}^{*}$, which in turn is used to update the learnable parameters $g$, $r$ and $h$. Note that the mixture feature vectors $\mathbf{c}$ and $\mathbf{c}^{*}$ and the weight $\mathbf{W}$ used here were also from the support set processing stage. 

During meta-training, each episode is presented one at a time, with the above-mentioned support set and query set processing repeated each time. In the proposed model structure, MetaCNN and the neural networks within the MergeNet and TconNet modules attempt to learn a good inductive bias through episodic meta-training, while the TAR mixture weight vectors $\boldsymbol{\omega}_{\text{pre}}$ and $\boldsymbol{\omega}_{\text{meta}}$ and the classification weights $\mathbf{W}^{*}$, which are generated anew given the support set of each new episode, provide quick task-conditioning for inference. Details of the training procedures and module operations are given in the following subsections.

\subsection{Training Procedures}


\textbf{Dataset Splits:} $\mathcal{D}_{\text{base}}$ denotes the dataset used in pretraining, and $\mathcal{D}_{\text{base/train}}$, $\mathcal{D}_{\text{base/val}}$ and $\mathcal{D}_{\text{base/test}}$ are training, validation and test sets of $\mathcal{D}_{\text{base}}$, respectively.
For incremental setting, dataset $\mathcal{D}_{\text{novel}}$ is for providing novel categories. The dataset consists of three splits of $\mathcal{D}_{\text{novel/train}}$, $\mathcal{D}_{\text{novel/val}}$ and $\mathcal{D}_{\text{novel/test}}$, containing disjoint categories.

\textbf{Pretraining phase:} On data split $\mathcal{D}_{\text{base/train}}$ with a fixed set of $N_{b}$ base classes, 
we train the embedding network $f_{\theta}$ and the base classifier weights of $\mathbf{W}$.

\textbf{Meta-training phase:} 
Meta-training is done on two data splits $\mathcal{D}_{\text{base/train}}$ and $\mathcal{D}_{\text{novel/train}}$. 
For each episode, $N$ novel classes are randomly chosen from training set $\mathcal{D}_{\text{novel/train}}$. Then we relabel these categories with new labels from $N_{b}+1$ to $N_{b}+N$.
Also, $K$ per-class data samples are randomly chosen from the set of novel categories for constructing a support set $\mathcal{S}=\{(\mathbf{x}_{j}^{S},y_{j}^{S})\}_{j=1}^{N\times K}$. 
The query set $\mathcal{Q}$ consists of samples not only from the novel but also the base categories.
To construct a novel query set $\mathcal{Q}_{\text{novel}}$, additional samples are randomly chosen from the novel categories.
 For base query samples, 
we randomly select the samples of base categories with the same size of $\mathcal{Q}_{\text{novel}}$ (i.e. $N\times K$) and then construct a base query set $\mathcal{Q}_{\text{base}}$.
The overall query set is constructed as a union of sets; $\mathcal{Q}=\mathcal{Q}_{\text{base}} \cup \mathcal{Q}_{\text{novel}}$. For each episode, learnable parameters are updated by computing loss during classification of queries in $\mathcal{Q}$.




\textbf{Test phase:} Upon deployment, all learnable parameters are fixed. 
For each test episode, $N$ novel categories are randomly chosen from split $\mathcal{D}_{\text{novel/test}}$ (or split $\mathcal{D}_{\text{novel/val}}$ for validation) which consists of unseen categories during both pretraining and meta-training phases.
For base categories, test samples are chosen from test split $\mathcal{D}_{\text{base/test}}$ (or split $\mathcal{D}_{\text{base/val}}$ for validation).



\subsection{MetaCNN for Extracting Novel Feature}
For a given input $\mathbf{x}$ from the support set, let $a_{\theta}(\mathbf{x})$ denote the intermediate layer output of the pretrained backbone $f_{\theta}$ and be used as an input to MetaCNN. This way, MetaCNN extracts a different high-level feature vector $g(a_{\theta}(\mathbf{x}))$. 
Let us call the vectors $f_{\theta}(\mathbf{x})\in\mathds{R}^{D}$ and $g(a_{\theta}(\mathbf{x}))\in\mathds{R}^{D}$ the base feature and novel feature, respectively. $D$ is the length of the features.



\subsection{MergeNet for Combining Base and Novel Features}
MergeNet $r$ is a meta-trained module for obtaining task-specific mixture weight vectors $\boldsymbol{\omega}_{\text{pre}}$ and $\boldsymbol{\omega}_{\text{meta}}$, which are used to combine the base and novel feature vectors. 
For each support sample in a given $\mathcal{S}$, the base and novel features are extracted by the pretrained backbone and MetaCNN; they are then concatenated. Let $[f_{\theta}(\mathbf{x}),g(a_{\theta}(\mathbf{x}))]$ denote this concatenated feature vector.
Afterwards, a task-representation vector $\mathbf{c}$ is computed by averaging all concatenated vectors of the support examples, i.e., $\mathbf{c}=\frac{1}{\lvert \mathcal{S} \rvert}\sum_{\mathbf{x}\in \mathcal{S}}[f_{\theta}(\mathbf{x}),g(a_{\theta}(\mathbf{x}))]$. 
MergeNet consists of two separate fully-connected networks $r_{\text{pre}}$ and $r_{\text{meta}}$, which compute $\boldsymbol{\omega}_{\text{pre}}$ and $\boldsymbol{\omega}_{\text{meta}}$ from $\mathbf{c}$, respectively: $\boldsymbol{\omega}_{\text{pre}} = r_{\text{pre}}(\mathbf{c})$ and $\boldsymbol{\omega}_{\text{meta}} = r_{\text{meta}}(\mathbf{c})$. 

After preparing $\boldsymbol{\omega}_{\text{pre}}$ and $\boldsymbol{\omega}_{\text{meta}}$, a combined feature vector is obtained by the summation of element-wise products between these mixture weight vectors and feature vectors:
$\boldsymbol{\omega}_{\text{pre}}\odot f_{\theta}(\mathbf{x}) + \boldsymbol{\omega}_{\text{meta}}\odot g(a_{\theta}(\mathbf{x}))$, where $\odot$ is element-wise product operation.
For each novel class $k$ in the given support set, the per-class average $\mathbf{c}_{k}^{*}$ of the combine feature vectors is calculated as
\begin{equation}
\mathbf{c}_{k}^{*} = \frac{1}{\lvert \mathcal{S}_{k} \rvert}\sum_{\mathbf{x}\in\mathcal{S}_{k}}[\boldsymbol{\omega}_{\text{pre}}\odot f_{\theta}(\mathbf{x}) + \boldsymbol{\omega}_{\text{meta}}\odot g(a_{\theta}(\mathbf{x}))],
\end{equation}
where $N_{b}+1\leq k \leq N_{b}+N$ and $\mathcal{S}_{k}$ is the set of support samples from novel class $k$.
For each episode, MergeNet learns to build the mixture weights which are used to construct the TAR.

\subsection{TconNet for Conditioning Classifiers}
TconNet $h$ consists of two meta-learned modules for task-conditioning of the classifier.
One module conditions the pretrained base classification weights $\{\mathbf{w}_{i}\}_{i=1}^{N_{b}}\in \mathbf{W}$.
First, the average of the combined feature vectors of support samples is obtained: $\mathbf{c}^{*}=\frac{1}{N}\sum_{k=N_b+1}^{N_b+N}\mathbf{c}_{k}^{*}$.
With two separate fully-connected networks $h_{\gamma}$ and $h_{\beta}$, two conditioning vectors are obtained from $\mathbf{c}^{*}$: $h_{\gamma}(\mathbf{c}^{*})$ and $h_{\beta}(\mathbf{c}^{*})$.
The task-conditioned base classification weights $\{\mathbf{w}_{i}^{*}\}_{i=1}^{N_{b}}$ are obtained by utilizing these vectors for scaling and biasing:
\begin{equation}
\mathbf{w}_{i}^{*}=(\mathds{1} + h_{\gamma}(\mathbf{c}^{*}))\odot\mathbf{w}_{i} + h_{\beta}(\mathbf{c}^{*})
\end{equation}
where $\mathds{1}$ is an all 1's vector with length $D$. The weights $\{\mathbf{w}_{i}^{*}\}_{i=1}^{N_{b}}$ are used as the base classifier.
We now discuss the initial classification weights 
 $\{\mathbf{w}_{i}\}_{i=N_b+1}^{N_b+N}\in\mathbf{W}$ for the novel categories. 

 \textbf{Preparing novel classifier:} Perhaps the simplest novel classifier setting is that of the Imprint method of \cite{imprint}: simply set the per-class feature averages as the novel classification weights, i.e., $\{\mathbf{c}_{k}^{*}\}_{k=N_b+1}^{N_b+N}$ is set to be $\{\mathbf{w}_{i}\}_{i=N_b+1}^{N_b+N}$. 
The initial novel classification weights can also be computed by LwoF, 
the attention-based method of \cite{dynamic}. 
This approach employs a meta-learned attention-based weight generator for the novel classifier.
When used with our XtarNet, the weight generator is learned along with other modules within XtarNet during meta-training. 
Yet another way is to utilize TapNet of \cite{TapNet}, which is a relatively simple but effective few-shot learning method. 
TapNet utilizes meta-learned per-class reference vectors for classification in a task-adaptive projection space. 
The reference vectors $\mathbf{\Phi}=\{\boldsymbol{\phi}_{n}\}_{n=1}^{N}$ of TapNet are learned across episodes while the projection space $\mathbf{M}$ is computed anew specific to each episode in such as way that class prototypes and the matching reference vectors align in $\mathbf{M}$. 
After obtaining per-class averages of embedded features, $\{\mathbf{c}_{n}\}_{n=1}^{N}$, space $\mathbf{M}$ is constructed such that 
$\{\boldsymbol{\phi}_{n}\}_{n=1}^{N}$ and $\{\mathbf{c}_{n}\}_{n=1}^{N}$ align in $\mathbf{M}$. More specifically, the errors defined as  
\begin{equation}
\boldsymbol{\epsilon}_{n}=\frac{\boldsymbol{\phi}_{n}}{\lVert\boldsymbol{\phi}_{n}\rVert} - \frac{\mathbf{c}_{n}}{\lVert\mathbf{c}_{n}\rVert}
\end{equation}
are forced to zero when projected in $\mathbf{M}$, i.e., $\mathbf{\boldsymbol{\epsilon}}_{n}\mathbf{M}=\mathbf{0}$ for all class $n$. 
Such an $\mathbf{M}$ can be found by 
taking the null-space of the errors by using singular value decomposition (SVD). 
When XtarNet is used in conjunction with TapNet, we simply set $\{\mathbf{w}_{i}\}_{i=N_b+1}^{N_b+N}\leftarrow \{\boldsymbol{\phi}_{n}\}_{n=1}^{N}$ and also utilize task-adaptive projection to perform classification. 

\textbf{Adaptation of novel classifier:} After the initial novel classifier $\{\mathbf{w}_{i}\}_{i=N_b+1}^{N_b+N}$ is obtained, another fully-connected network $h_{\lambda}$ of TconNet learns to separate novel classifier weights from the task-conditioned base classifier weights by introducing bias terms.
First, a correlation vector $\boldsymbol{\sigma}_{k}$ is calculated for each novel class $k$, whose $i^{\text{th}}$ element $\sigma_{k}^{i}$ is the inner-product similarity between the combined per-class average feature $\mathbf{c}_{k}^{*}$ and the base classification weight, i.e., $\sigma_{k}^{i}=\mathbf{c}_{k}^{*}\cdot\mathbf{w}_{i}^{*}$ for all $i=1,\cdots,N_b$. 
For each novel class $k$, the module takes the correlation vector $\boldsymbol{\sigma}_{k}\in\mathbb{R}^{N_{b}}$ as the input to calculate: $\boldsymbol{\lambda}_{k}=h_{\lambda}(\boldsymbol{\sigma}_{k})\in \mathbb{R}^{N_b}$.
Then the output vector is utilized to modify $\mathbf{w}_{i}$ by introducing a bias term, which is a weighted sum of the conditioned base classification weights $\{\mathbf{w}_{j}^{*}\}_{j=1}^{N_b}$:
\begin{equation}
\mathbf{w}_{i}^{*}=\mathbf{w}_{i} - \displaystyle\sum_{j=1}^{N_b}\frac{\exp(\lambda_{i}^{j})}{\sum_{l}{\exp(\lambda_{i}^{l})}}\mathbf{w}_{j}^{*}
\end{equation}
for all $N_{b}+1\leq i \leq N_{b}+N$. $\lambda_{k}^{i}$ is $i^{\text{th}}$ element of $\boldsymbol{\lambda}_{k}$.
This completes the overall set of the task-conditioned weight vectors 
$\mathbf{W}^{*}=\{\mathbf{w}_{i}^{*}\}_{i=1}^{N_b+N}$ for classifying queries.

For best results, our XtarNet utilizes the task-adaptive projection space of TapNet. The projection space $\mathbf{M}$ is computed such that the per-class averages of the combined features $\{\mathbf{c}_{k}^{*}\}_{k=N_b+1}^{N_b+N}$ and the conditioned novel classifier weights of $\mathbf{W}^{*}$ coincide in $\mathbf{M}$.

For a given query $\mathbf{x}\in\mathcal{Q}$, the Euclidean distance $D_{i}(\mathbf{x})$ between the projected feature of query and the projected classifier weights is used in classification:
\begin{equation}\label{eq:dist}\small
D_{i}(\mathbf{x})=d\Big(\mathbf{w}_{i}^{*}\mathbf{M}, [\boldsymbol{\omega}_{\text{pre}}\odot f_{\theta}(\mathbf{x}) + \boldsymbol{\omega}_{\text{meta}}\odot g(a_{\theta}(\mathbf{x}))]\mathbf{M} \Big),
\end{equation}
Finally, classification of the query $\mathbf{x}$ is carried out based on the posterior probabilities:
\begin{equation}
p(y=i|\mathbf{x})=\frac{ \exp (-D_{i}(\mathbf{x}) )  }
{\sum_{l} \exp (-D_{l}(\mathbf{x}) )  },
\end{equation}
for all $i=1,\cdots,N_{b}+N$. 

A pseudocode-style algorithm description of XtarNet is given in Supplementary Material.

\section{Related Work}
\textbf{Few-shot learning:}
Few-shot learning attempts to train models for novel tasks with only a few labeled samples. 
The metric-learning-based approach is one of the most popular branches of few-shot learning. The key idea of metric-based few-shot learners is to train the feature embedding so that the embedded samples belonging to the same class are located closely. 
Matching Networks of \cite{MN} learn the embedding space and the query samples are classified based on the similarity scores computed by the distance between the few labeled support samples and query samples. 
Note that two different embedding spaces can be learned to process the support set and query set.    
Prototypical Networks of \cite{PN} also rely on the embedding network where the samples from the same class are clustered together. Instead of computing similarity between samples, Prototypical Networks utilize per-class averages as the prototypes. Classification of queries is done by finding the nearest prototype in the embedding space. 
	
\textbf{Task-conditioning:}
Recently, few-shot learning algorithms utilize explicit task-conditioning and achieve significant performance gain. TADAM of \cite{TADAM} offers a task-conditioning method based on the general-purpose conditioning of FiLM proposed in \cite{FILM}.
Like FiLM, TADAM performs affine transformation on the middle level feature. The scaling and shifting parameters are generated from task representation, which is the averaged feature of support samples. An additional task-conditioning module is meta-learned to generate the conditioning parameters, and auxiliary co-training is adopted for efficient meta-learning. The component module TconNet of our XtarNet also provides task-conditioning in a form of affine transformation to adjust the base classifier. 
TapNet of \cite{TapNet} proposed another type of task-conditioning based on feature projection. Additional meta-learned reference vectors are utilized in classification. When a task is given, a projection space is constructed such that the class prototypes and the reference vector closely align there. Classification is done in the projection space by measuring distance between projected queries and references. 
These prior works and our method are common in the sense that explicit task-conditioning is employed, but our XtarNet is unique in that task-conditioning is made possible by creating a mixture of the base and novel features that depends on the given task.

       
\textbf{Incremental learning:}
Learning from the continuously arriving data while preserving learned knowledge is called incremental learning. The key to incremental learning is preventing catastrophic forgetting \cite{CF_1, CF_2}, which refers to the situation where the model forgets learned prior knowledge. 
One way to prevent catastrophic forgetting is using memory \cite{inc_mem_1,inc_mem_3}. The memory-based model stores prior training data explicitly and use it for training again at the appropriate moment. Generative models can also be used to prevent forgetting by generating previously learned data when data is needed \cite{inc_gen_1, inc_gen_2}. 
Regularization can be used for learning from new data while preventing catastrophic forgetting, without replaying previous data \cite{inc_reg_1}.
Learning without forgetting is another way to prevent forgetting \cite{inc_lwof}. The information about prior data is not stored, and feature extracted from new data using an old model is utilized to preserve prior knowledge.
Our XtarNet is closest to the Learning without forgetting method of \cite{inc_lwof}, in terms of utilizing new data from novel categories to prevent forgetting of base categories.

		
\textbf{Incremental few-shot learning:}
The objective of incremental few-shot learning is to obtain an ability to classify novel classes while preserving the capability to classify base classes, when only a few labeled samples are given for novel classes. 
In general, the backbone network and classification weights for base classes are pretrained by regular supervised learning and are fixed afterwards. To our best knowledge, all prior works on incremental few-shot learning focus on generating the classification weights for novel classes. 
The Weights Imprinting method (Imprint) of \cite{imprint} computes the prototypes of novel classes and use them as classification weights for novel classes. Another method, 
Learning without Forgetting (LwoF) of \cite{dynamic} learns to generate the novel classification weights by a meta-learned weight generator that takes novel prototypes and base weights as inputs. LwoF utilizes an attention-based mechanism to exploit the relation between base classes and novel classes in the generation of novel weights. Yet another prior work, Attention Attraction Network of \cite{incMeta}, trains the novel classification weights by a gradient-based optimization process using cross entropy loss from a few labeled samples of the novel classes until they converge. Since the loss for training novel weights is computed only with the samples of novel classes, the catastrophic forgetting issue for base classes may arise. 
To prevent this, the attention-based regularization method is applied. The regularization loss is provided by a meta-learned attention attractor network. The attention attractor network generates attractor vectors using the base classification weights, and regularization loss is computed based on the Mahalanobis distances between the novel classification weights and attractor vectors.
These prior works on incremental few-shot learning focus only on learning to construct novel classifiers.
In contrast, our XtarNet aims to extract novel representation which cannot be provided by the pretrained backbone alone.

\begin{table*}[h]
\centering
\begin{threeparttable}
	\caption{\textit{mini}ImageNet 64+5-way results}
	\label{table:mini}
	\begin{tabular}{l||cc||cc}
		\toprule  
		& \multicolumn{2}{c}{\textbf{1-shot}} & \multicolumn{2}{c}{\textbf{5-shot}} \\
		\cmidrule{2-5}
		\textbf{Methods}   & Accuracy & $\Delta$ 	& Accuracy  & $\Delta$  \\
		\midrule
		\textbf{Imprint} \cite{imprint}  &  41.34 $\pm$ 0.54\%	& -23.79\% & 46.34 $\pm$ 0.54\%	& -25.25\%  \\
		\textbf{LwoF} \cite{dynamic} &  49.65 $\pm$ 0.64\% 	& -14.47\% & 59.66 $\pm$ 0.55\% & -12.35\%  \\
		\textbf{Attention Attractor Networks}* \cite{incMeta} & 54.95 $\pm$ 0.30\% 	& \textbf{-11.84}\% & 63.04 $\pm$ 0.30\%	& -10.66\%  \\
		\textbf{XtarNet} (Ours, utilizing TapNet)  &  \textbf{55.28} $\pm$ \textbf{0.33}\%	& -13.13\% & \textbf{66.86} $\pm$ \textbf{0.31}\%	& \textbf{-10.34}\%  \\
		\midrule
		\textbf{Proposed method with Imprint}  &  47.96 $\pm$ 0.59\%	& -12.72\% & 56.66 $\pm$ 0.51\%	& -11.23\%  \\
		\textbf{Proposed method with LwoF}  &  49.71 $\pm$ 0.61\%	& -13.54\% & 60.13 $\pm$ 0.50\%	& -11.40\%  \\
		\bottomrule
		\end{tabular}
		
		\end{threeparttable}
\end{table*}
\begin{table*}[h]
\centering
\begin{threeparttable}
	\caption{\textit{tiered}ImageNet 200+5-way results}
	\label{table:tiered}
	\begin{tabular}{l||cc||cc}
		\toprule  
		& \multicolumn{2}{c}{\textbf{1-shot}} & \multicolumn{2}{c}{\textbf{5-shot}} \\
		\cmidrule{2-5}
		\textbf{Methods}   & Accuracy & $\Delta$ 	& Accuracy  & $\Delta$  \\
		\midrule
		\textbf{Imprint} \cite{imprint}  &  40.83 $\pm$ 0.45\%	& -22.29\% & 53.87 $\pm$ 0.48\%	& -17.18\%  \\
		\textbf{LwoF} \cite{dynamic} &  53.42 $\pm$ 0.56\% 	& -9.59\% & 63.22 $\pm$ 0.52\% & -7.27\%  \\
		\textbf{Attention Attractor Networks}* \cite{incMeta} & 56.11 $\pm$ 0.33\% 	& -6.11\% & 65.52 $\pm$ 0.31\%	& -4.48\%  \\
		\textbf{XtarNet} (Ours, utilizing TapNet)  &  \textbf{61.37} $\pm$ \textbf{0.36}\%	& \textbf{-1.85}\% & \textbf{69.58} $\pm$ \textbf{0.32}\%	& \textbf{-1.79}\%  \\
		\midrule
		\textbf{Proposed method with Imprint}  &  56.37 $\pm$ 0.56\%	& -8.89\% & 65.56 $\pm$ 0.50\%	& -6.82\%  \\
		\textbf{Proposed method with LwoF}  &  55.41 $\pm$ 0.57\%	& -9.33\% & 65.68 $\pm$ 0.51\%	& -6.68\%  \\
		\bottomrule

	\end{tabular}
	\begin{tablenotes}   
 		\item *The Attention Attractor Networks results reflect the reported numbers of \cite{incMeta}.
 		\end{tablenotes}
		\end{threeparttable}
\end{table*}

\section{Experimental Result}
We follow the settings of incremental few-shot classification tasks proposed in \cite{dynamic, incMeta} which are based on the \textit{mini}ImageNet and \textit{tiered}ImageNet datasets. 
Pretraining is done with $\mathcal{D}_{\text{base}}$ consisting of 64 and 200 base categories for \textit{mini}ImageNet and \textit{tiered}ImageNet experiments, respectively.
In the meta-training phase, 5 novel categories are selected from $\mathcal{D}_{\text{novel/train}}$ for each episode. For \textit{tiered}ImageNet experiments, $\mathcal{D}_{\text{novel/train}}$ contains additional 151 categories. 
Following the setting of \cite{dynamic, incMeta}, for \textit{mini}ImageNet experiments, $\mathcal{D}_{\text{base/train}}$ is reused as $\mathcal{D}_{\text{novel/train}}$; 5 base categories are selected as fake novel categories for each episode. For deployment, $\mathcal{D}_{\text{novel/val}}$ and $\mathcal{D}_{\text{novel/test}}$ serve as the validation and test sets for each dataset, respectively.

For each episode, the support set is constructed by choosing 1 or 5 examples (or shots) from 5 randomly selected novel categories. The query set consists of samples from both base and novel categories.
For the meta-training phase, we optimize the number of per-class queries for novel categories. 

For the \textit{mini}ImageNet experiments, we use ResNet12 adopted in \cite{SNAIL} as the backbone architecture. 
MetaCNN consists of a single residual block with the same size as the last residual block of the backbone. Also, MetaCNN takes the output of the third residual block of the backbone as its input. MergeNet contains two separate 4-layer fully-connected networks. For TconNet, two separate 3-layer fully-connected networks with a skip connection for each layer are used to condition the base classifier.
Finally, an additional 3-layer fully-connected network with a skip connection for the last two layers is used to condition the novel classifier.
In the meta-learning phase, the MomentumSGD optimizer with an initial learning rate of $0.1$ is utilized. The learning rate is dropped by a factor of $10$ at every step of $4,000$ episodes. For regularization in meta-learning, $l2$ regularization with the ratio $3.0\times10^{-3}$ is used.


For the \textit{tiered}ImageNet experiments, we use the standard ResNet18 architecture. 
MetaCNN contains two residual blocks with the same size as the last two blocks of the backbone. Other modules are based on the same architectures that are used in the \textit{mini}ImageNet experiments.
In the meta-learning phase, the MomentumSGD optimizer with an initial learning rate of $0.1$ is utilized. The learning rate is dropped by a factor of $10$ at every step of $4,000$ episodes. For regularization in the meta-learning, $l2$ regularization with the ratio $7.0\times10^{-4}$ is used.



\subsection{Evaluation Metrics}
We can evaluate the proposed model using joint accuracy, which is evaluated using query sets sampled jointly from base and novel classes. The joint accuracy indicates the capability of models to handle base and novel queries together.
Also, the gaps between joint accuracy and individual accuracies for base and novel classes are evaluated. Individual accuracy for base (or novel) classes are obtained by classifying the $\mathcal{Q}_{\text{base}}$ (or $\mathcal{Q}_{\text{novel}})$ query set using only the base (or novel) classifier. 
The gaps for base and novel classes are denoted by $\Delta_{a}$ and $\Delta_{b}$, respectively.
The average of these gaps $\Delta = \frac{1}{2}(\Delta_{a}+\Delta_{b})$ is evaluated. 

\begin{table*}[h]
	\caption{Ablation studies for the 5-shot \textit{tiered}ImageNet case}
	\centering
	\label{table:ablation}
	\begin{tabular}{l||cc}
		\toprule  
		& \multicolumn{2}{c}{\textbf{5-shot}} \\
		\cmidrule{2-3}
		\textbf{Methods}   & Accuracy  & $\Delta$ 	\\
		\midrule
		\textbf{Imprint} \cite{imprint}  & 53.87 $\pm$ 0.48\%	& -17.18\%  \\
		\midrule
		\textbf{MetaCNN with Imprint}  & 63.15 $\pm$ 0.49\%	& -7.32\%  \\
		\textbf{MetaCNN + MergeNet with Imprint} & 64.62 $\pm$ 0.51\%	& -7.09\%  \\
		\textbf{MetaCNN + MergeNet + TconNet with Imprint} & 65.56 $\pm$ 0.50\%	& -6.82\%  \\
		\bottomrule
	\end{tabular}
\end{table*}

\begin{figure*}[h]
\centering
\begin{subfigure}[b]{0.33\textwidth}
\includegraphics[width=\textwidth]{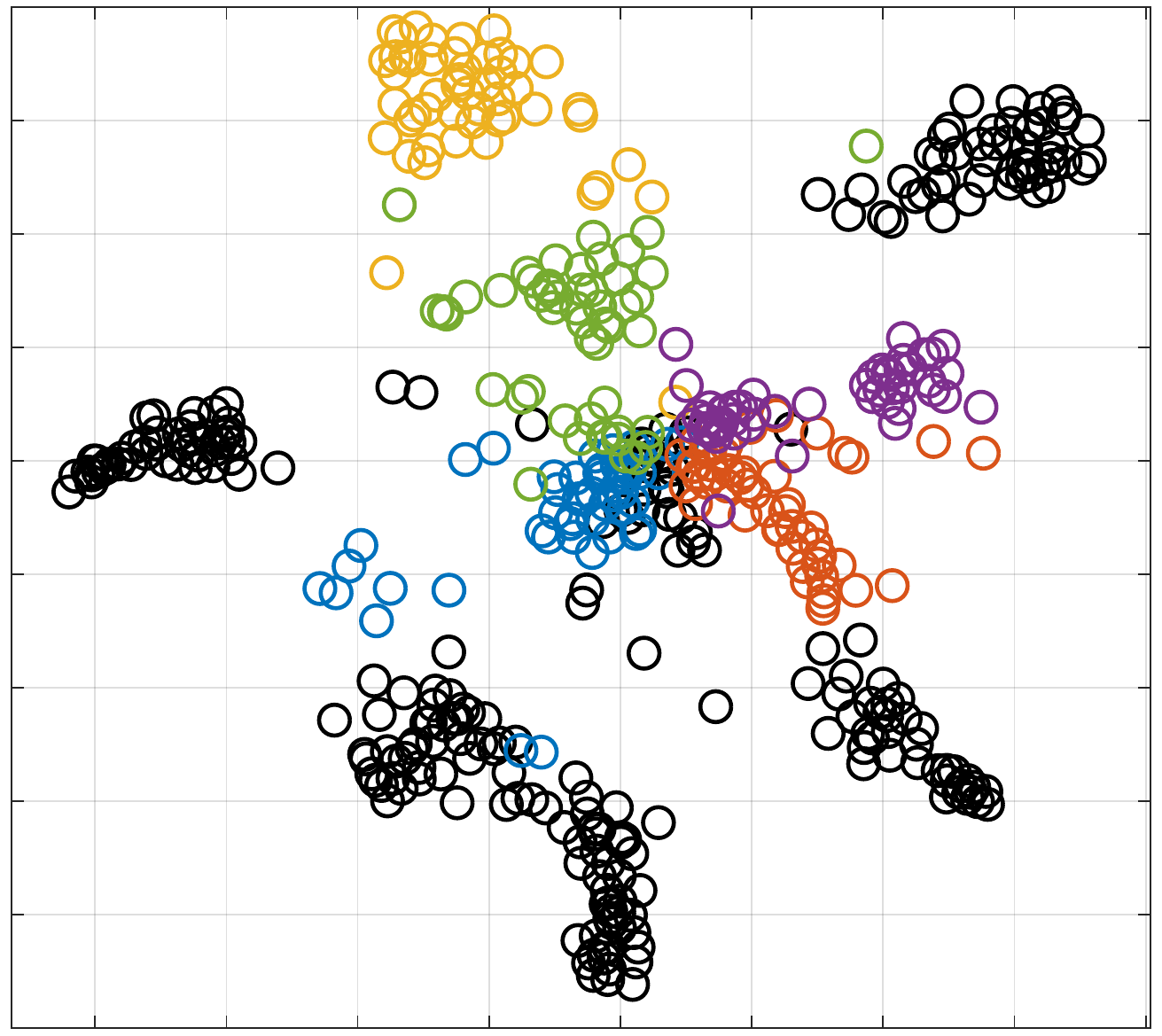}
\subcaption{Features from the pretrained backbone}
\label{fig:tsne1}
\end{subfigure}
\begin{subfigure}[b]{0.33\textwidth}
\includegraphics[width=\textwidth]{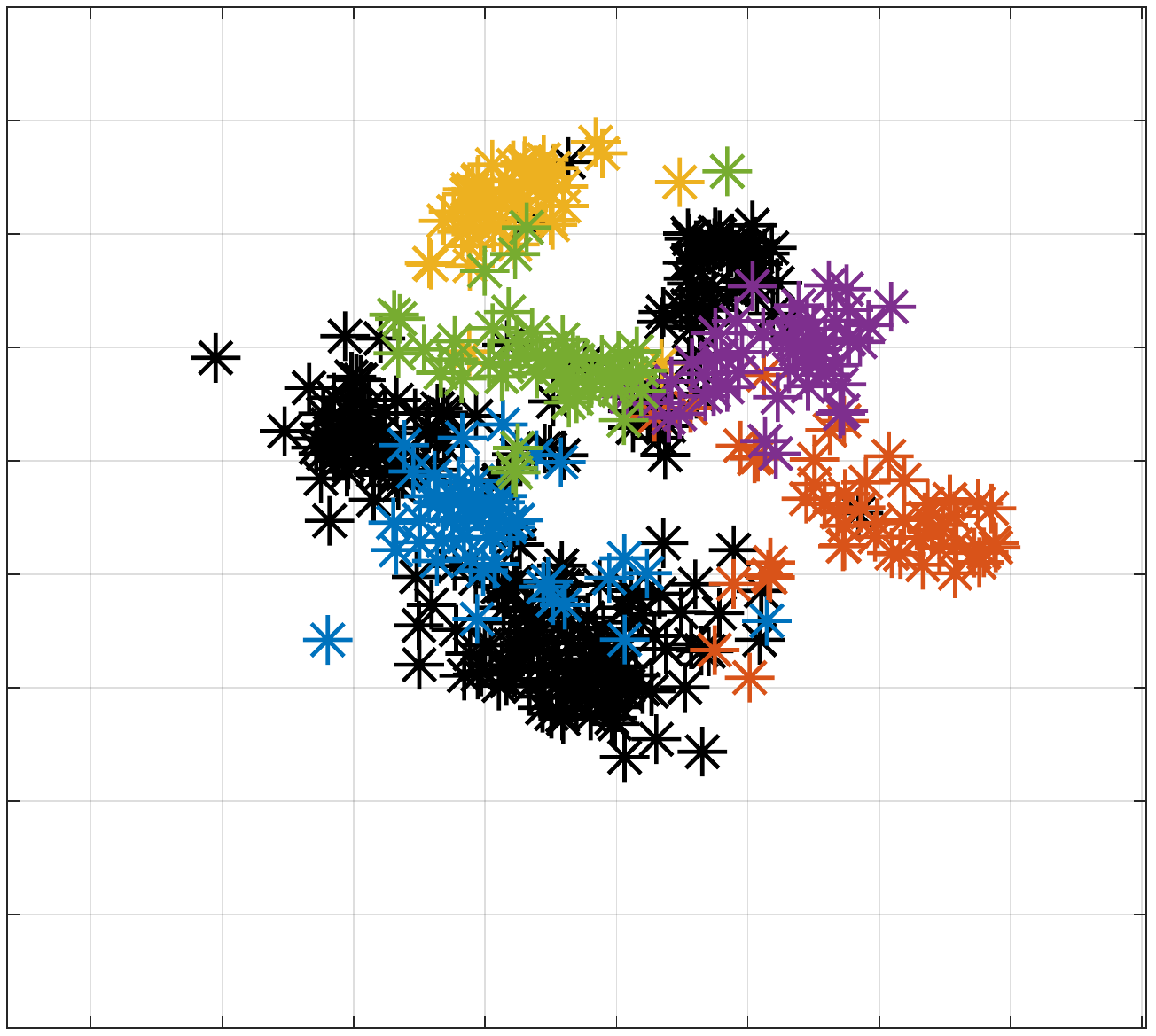}
\subcaption{Features from MetaCNN}
\label{fig:tsne2}
\end{subfigure}
\begin{subfigure}[b]{0.33\textwidth}
\includegraphics[width=\textwidth]{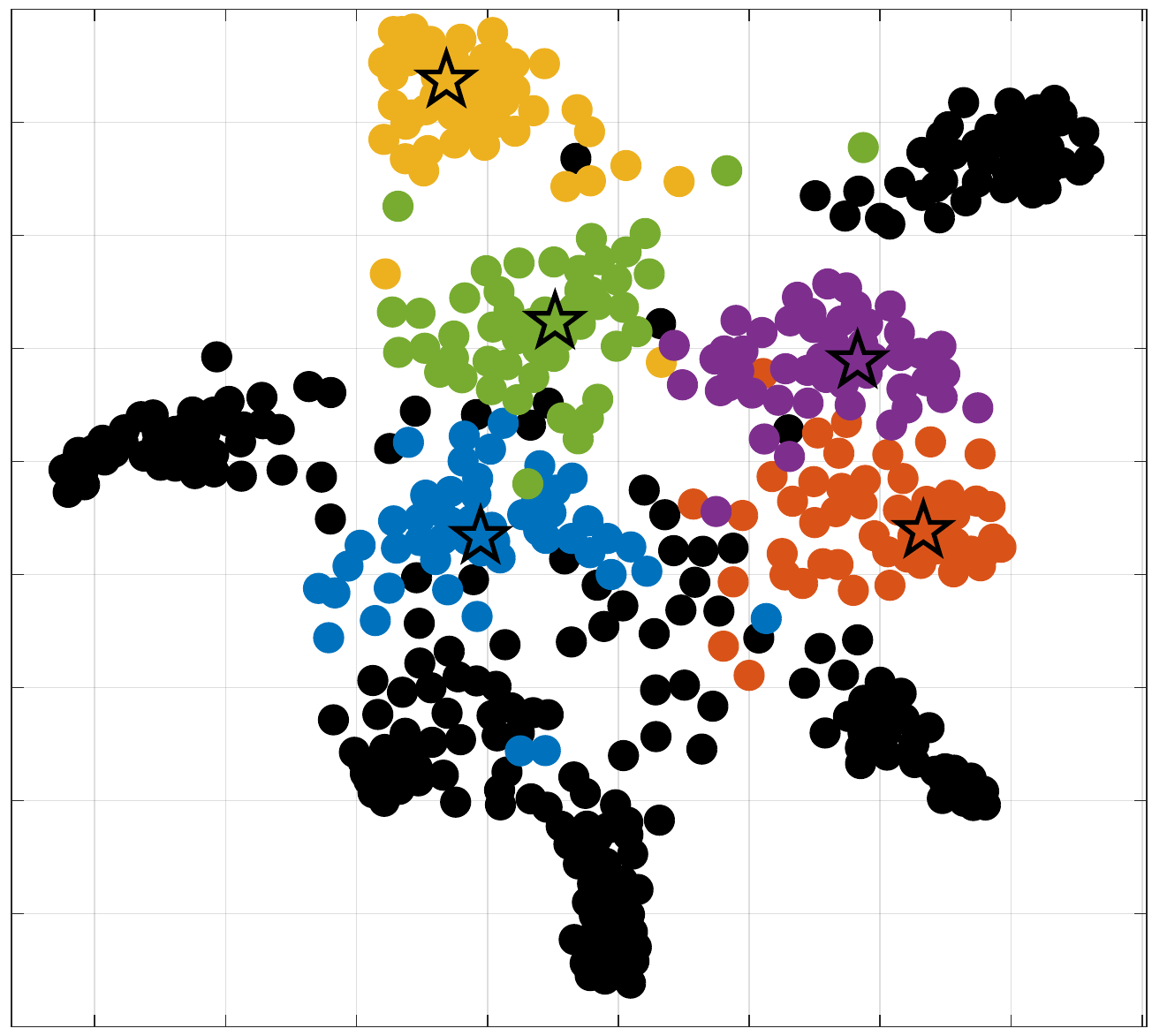}
\subcaption{Combined features}
\label{fig:tsne3}
\end{subfigure}
\caption{tSNE plots of queries from a test episode for the proposed TAR method built on Imprint}
\label{fig:tsne}
\end{figure*}

\subsection{Similarity Metrics}
For XtarNet, we use the Euclidean distance metric in classification. For the pretraining, the backbone network is also trained with the Euclidean distance metric.
We evaluate other two versions of our proposed method with relatively simple prior methods of Imprint and LwoF, respectively. 
We use cosine similarity metric for both methods with Imprint and LwoF.
For these cases, a backbone is also pretrained with the cosine similarity metric.

\subsection{Results}
In Tables \ref{table:mini} and \ref{table:tiered}, we present the results of incremental few-shot learning experiments with the \textit{mini}ImageNet and \textit{tiered}ImageNet datasets\footnote{Codes are available on \url{https://github.com/EdwinKim3069/XtarNet}}. All results for prior methods except for Attention Attractor Networks are reproduced here. The Attention Attractor Networks results shown in the tables are the published numbers of \cite{incMeta}. For our XtarNet, we used TapNet for creating the initial novel classifier. Also, the TapNet projection space is utilized in running XtarNet.
The performance is evaluated by the averaged accuracy with a 95\% confidence interval over $2,000$ test episodes. The $\Delta$ values are also evaluated.
For each test episode, we use a query set containing the samples of size $25$ for novel classes and $25$ for base classes.

For the 5-shot \textit{mini}ImageNet case, XtarNet shows the best joint accuracy and $\Delta$ value. For the 1 
and 5-shot \textit{tiered}ImageNet cases, XtarNet achieves the best performances with considerable margins in terms of both joint accuracies and $\Delta$ values.
For the 1-shot \textit{mini}ImageNet case, on the other hand, XtarNet shows the second best performance in $\Delta$ value. 

Moreover, the proposed methods in conjunction with Imprint and LwoF also yield significant performance advantages over the base Imprint and LwoF methods. 
In particular, for Imprint, which shows severely degraded performance for incremental few-shot learning tasks, remarkable gains are shown for all test cases when combined with the propose TAR.
For the 1 and 5-shot \textit{tiered}ImageNet tests, the proposed idea combined with Imprint shows even better accuracies than existing incremental few-shot learning algorithms. Plugging into Attention Attractor Network of \cite{incMeta} is also possible and the results are presented in Supplementary Material.


\subsection{Ablation Studies}
Table \ref{table:ablation} shows the evaluation results of one-by-one employment of our meta-learned modules MetaCNN, MergeNet and TconNet. 
Our ablation studies are done with the Imprint method which does not contain any meta-trained modules for preparing novel classifier weights, so that the results highlight the gains obtained by applying the concept of TAR.
Adding just MetaCNN yields a large gain over the baseline Imprint method. 
Adopting MergeNet gives another 1.5\% or so gain; adding the TconNet module finally yields yet another 0.9\% gain, brining the combined gain of 
more than 11\% overall. 

\section{Analysis of Task-Adaptive Representation}
\subsection{tSNE Analysis on Base and Novel Features} In Fig. \ref{fig:tsne}, tSNE analysis is presented for illustrating how MetaCNN extracts novel features and how MergeNet combines them with base features to acquire effective combined features.
The proposed method with Imprint is considered for highlighting gains obtained by employing the concept of TAR.
We consider the 5-shot \textit{tiered}ImageNet incremental few-shot classification tests.
For the analysis, features of 50 per-class queries from a test episode are considered.
In the three subfigures of Fig. \ref{fig:tsne}, the black-colored symbols represent base class queries from five selected base categories, and the five colored symbols indicate queries from five novel categories.
In Fig. \ref{fig:tsne1}, the weighted base feature vectors $\boldsymbol{\omega}_{\text{pre}}\odot f_{\theta}(\mathbf{x})$ of queries are presented with a circle symbol $\boldsymbol{\circ}$.
The queries from base classes, the black symbols, are clustered and show good separation.
The good clustering and separation make sense because the backbone is pretrained to classify base classes. However, the weighted base feature vectors of queries from novel classes do not show good clustering behavior. 
In particular, the four novel classes depicted with green, blue, purple and orange colors are severely stuck together and indistinguishable. In Fig. \ref{fig:tsne2}, the weighted novel feature vectors of queries from MetaCNN are illustrated with the star symbol $\boldsymbol{\ast}$. In comparison with the base feature vectors, MetaCNN clearly provides improved features for identifying novel classes. 
 Finally, the combined feature vectors are presented with the bullet symbol $\boldsymbol{\bullet}$ in Fig. \ref{fig:tsne3}.
For base class queries, they remain well-clustered as in Fig \ref{fig:tsne1}. 
On the other hand, the combine feature vectors of novel classes have changed dramatically. By adopting the features from MetaCNN, the combined features are reasonably clustered in comparison with the base features depicted in Fig. \ref{fig:tsne1}. The per-class averages of the combined feature vectors are presented with the symbols $\medstar$ which work as classification weights of novel classes for the proposed method with Imprint. 
The per-class averages of novel classes become more separated so that they can provide an appropriate novel classifier. This is the reason why adopting our methods show considerable improvement over the baseline Imprint method.

\subsection{Quality of Clustering}
Let us take a closer, quantitative look at how the proposed TAR idea improves upon Imprint. 
To measure the quality of clustering for each category, the sum of squared distances or errors (SSE) between embedded queries and their centroid is evaluated. The SSE is calculated as follows:
\begin{equation}\small
\text{SSE}_{k}=\sum_{\mathbf{x}\in\mathcal{Q}_{k}} \big\lvert\big\lvert \mathbf{c}_{k}^{*} - \big(\boldsymbol{\omega}_{\text{pre}}\odot f_{\theta}(\mathbf{x}) + \boldsymbol{\omega}_{\text{meta}}\odot g(a_{\theta}(\mathbf{x}))\big) \big\rvert\big\rvert^{2}
\end{equation}
where $\mathcal{Q}_{k}$ is a set of query samples from class $k$.
Meanwhile, the SSE of the Imprint method is calculated as:
\vspace{-0.05cm}
\begin{equation}\small
\text{SSE}_{k}=\sum_{\mathbf{x}\in\mathcal{Q}_{k}}\big\lvert\big\lvert \mathbf{c}_{k} - f_{\theta}(\mathbf{x})\big\rvert\big\rvert^{2}.
\end{equation}

For each test episode for the 5-shot 200+5 \textit{tiered}ImageNet classification, the SSE values for each method are averaged for base and novel categories. 50 queries are considered for each class. Results are averaged over 600 episodes. Table \ref{table:SSE} shows the averaged SSE values of Imprint as well as the combined method.
Let us focus on the first column for each base and novel case in Table \ref{table:SSE}.
It is clearly shown that applying TAR significantly improves the quality of clustering for both base and novel categories. 
From the SSE analysis, it is apparent that the TAR enhances the quality of clustering for both base and novel categories.

To consider the interference from adjacent clusters, SSE values can be normalized by the squared distance to the nearest interfering centroid (denoted by nSSE). The results are reported in the right column of each base and novel case in Table \ref{table:SSE}.
When the inference is considered, TAR shows much stronger effect on novel classes.
\begin{table}
\small
	\caption{Clustering quality analysis for \textit{tiered}ImageNet results}
	\centering
	\label{table:SSE}
	\begin{tabular}{l||cc|cc}
		\toprule  
		& \multicolumn{2}{c|}{Base} & \multicolumn{2}{c}{Novel} \\
		\cmidrule{2-5}
		\textbf{Methods}   & SSE & nSSE & SSE & nSSE 	\\
		\midrule
		\textbf{Imprint}  & 10.65 &  81.02 & 3.96 & 250.31    \\
		\textbf{Combined}  & 4.94 & 76.92 & 2.38 & 179.51   \\
		\textbf{Reduction Ratio} & -53.6\% & -5.06\% & -39.9\% & -28.3\% \\
		\bottomrule
	\end{tabular}
\end{table}

\subsection{Entropy} 
Let us now conduct entropy analysis with two information-theoretic metrics: cross entropy $\mathcal{E}$ and Shannon entropy $\mathcal{H}$. 
We use the same setting as in the analysis for the quality of clustering.
For each test episode from \textit{tiered}ImageNet, 50 per-class queries are selected and the averages of their entropy values are evaluated.
For two groups of base and novel categories, the means of the averaged entropy values are calculated.
A smaller cross-entropy value means that inference on labels is more accurate. A smaller Shannon entropy value indicates that inference is more confident.
The results in Table \ref{table:entropy} shows that inference on test queries becomes more accurate and confident by means of TAR.
\begin{table}[h]
\small
	\caption{Entropy analysis for \textit{tiered}ImageNet results}
	\centering
	\label{table:entropy}
	\begin{tabular}{l||cc|cc}
		\toprule  
		& \multicolumn{2}{c|}{Base} & \multicolumn{2}{c}{Novel} \\
		\cmidrule{2-5}
		\textbf{Methods}   & $\mathcal{E}$ & $\mathcal{H}$ & $\mathcal{E}$ & $\mathcal{H}$ 	\\
		\midrule
		\textbf{Imprint}  & 4.78 &  5.31 	& 4.97 & 5.31    \\
        \textbf{Combined}  & 1.61 & 1.48 & 0.92 & 1.37   \\
		\textbf{Reduction Ratio} & -66.3\% & -72.1\% & -81.5\% & -74.2\% \\
		\bottomrule
	\end{tabular}
\end{table}


\section{Conclusions}
We proposed an incremental few-shot learning algorithm which learns to extract task-adaptive representation for classifying both base and novel categories. Built on a pretrained model, our proposed method XtarNet employs three essential modules that are meta-trained.
For a given new task, XtarNet extracts novel features which cannot be captured by the pretrained backbone. The novel feature is then combined with the base feature captured by the backbone. The mixture of these features constructs task-adaptive representation, facilitating incremental few-shot learning. 
The base and novel classifier quickly adapt to the given task by utilizing TAR. 
The concept of TAR can be used in conjunction with known incremental few-shot learning methods, and significant performance gains are achieved.
For the incremental few-shot learning benchmarks using \textit{mini}ImageNet and \textit{tiered}ImageNet, our XtarNet achieves state-of-the-art accuracies.

\section*{Acknowledgements}
This work was supported by IITP funds from MSIT of Korea (No. 2016-0-00563, No. 2020-0-00626 and No. 2020-0-01336) for KAIST and AI Graduate School Program at UNIST.


\bibliography{ICML2019}
\bibliographystyle{icml2019}


\section*{Supplementary material}

# SUPPLEMENTAL MATERIAL
# XtarNet: Learning to Extract Task-Adaptive Representation for Incremental Few-Shot Learning

**Sung Whan Yoon**[* 1]  **Do-Yeon Kim**[* 2]  **Jun Seo**[2]  **Jaekyun Moon**[2]

## 1. Details of Backbone Networks

For the *mini*ImageNet experiments of XtarNet, we utilized ResNet12, a residual network with 12 convolutional layers, as architecture of the backbone network. ResNet12 consists of four residual blocks having 64, 96, 128 and 256 respective channels. Each residual block contains three $3 \times 3$ convolutional layers, where every convolutional layer is followed by a batch normalization and an ReLU activation. For the last layer of the fourth residual block, we do not use an ReLu activation. The residual connection of each block consists of a single $3 \times 3$ convolutional layer and a batch normalization layer. At the end of each residual block, $2 \times 2$ max-pooling is applied. To reduce the dimension of the feature vector, a global average pooling is applied to the output of the last residual block. For every residual block, we use a dropout layer with a drop ratio of $0.2$ after the max-pooling layer. For all methods other than XtarNet, the only difference is that we use an ReLu activation at the last layer of the fourth residual block.

For the *tiered*ImageNet experiments, we employed the standard ResNet18, which consists of 18 convolutional layers, as the backbone network. As the *mini*ImageNet cases, we also do not use an ReLu activation for the last layer of the fourth residual block. ResNet18 is well-known, and we forgo a detailed description here. Also, for all other methods except XtarNet, we use an ReLu activation layer for every residual blocks.

## 2. Details of Meta-Trained Modules

The details of MetaCNN module are described in the main paper. We utilize fully-connected networks for MergeNet and TconNet. For the *mini*ImageNet experiments, MergeNet consists of two separate 4-layer fully-connected networks. These two component networks have the same architecture. The size of the input layer of each 4-layer fully-connected network is 512, which is the length of a concatenated feature vector. For the hidden layers and the output layer, the size of each layer is 256. For each of first three fully-connected layers, an ReLu activation is employed. For the last fully-connected layer, a sigmoid layer is employed. TconNet consists of three 3-layer fully-connected networks. Two of them are for computing the conditioning vector for the base classifier. These two fully-connected networks have a skip connection at each layer. The size of each fully-connected layer is 256, which is the length of a combined feature vector. ReLu activation is used at every layer. At each output of these two component fully-connected networks, we multiply the output vector by the constant 0.1 for scaling. The remaining 3-layer fully-connected network is for conditioning of the novel classifier. The size of each fully-connected layer is 64, which is the number of the base categories. The last two fully-connected layers have skip connections. ReLu activation is used at every layer.

For the *tiered*ImageNet experiments, the architectures of MergeNet and TconNet are the same as that we used in the *mini*ImageNet experiments, except for the size of the fully-connected layers. Because the length of the feature vector becomes double, the size of every fully-connected layer of MergeNet and TconNet also doubles. In TconNet, the size of each fully connetected layer for conditioning novel classifier is 200, which is the number of base categories of *tiered*ImageNet.

## 3. Hyperparameter Settings

We use $25 \times 2 = 50$ queries in each training episode for the proposed XtarNet algorirthm. In the *tiered*ImageNet experiments, we also use 50 queries for the TAR method used in conjunction with Imprint of (Qi et al., 2018), LwoF of (Gidaris & Komodakis, 2018) and Attention Attractor Network (AAN) of (Ren et al., 2019), respectively. On the other hand, we used $5 \times 5$ for base categories and $25 \times 5$

---

[*]Equal contribution  [1]School of Electrical and Computer Engineering, Ulsan National Institute of Science and Technology (UNIST), Ulsan, Korea  [2]School of Electrical Engineering, Korea Advanced Institute of Science and Technology (KAIST), Daejeon, Korea. Correspondence to: Sung Whan Yoon <shyoon8@unist.ac.kr>.

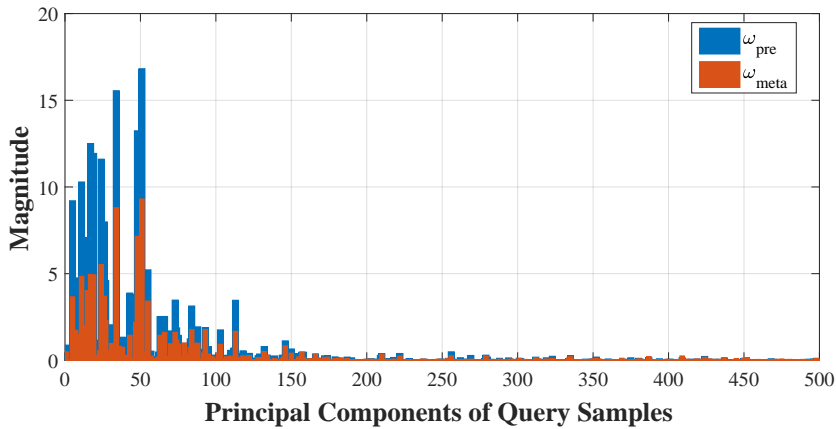

Figure 1. Principal component analysis for mixture weight vectors

for novel categories in TAR plugged-in Imprint and LwoF for the *mini*ImageNet experiments. For all cases, we use the Momentum SGD optimizer with an initial learning rate of 0.1. The learning rate decays by a factor of 10 at every step of 4,000 episodes. $l2$ weight decay is set to be $3.0 \times 10^{-3}$ for all *mini*ImageNet experiments and $7.0 \times 10^{-4}$ for all *tiered*ImageNet experiments. Also, the scaling hyperparameters $a$ and $b$ for outputs of meta-trained modules $h_\gamma$ and $h_\beta$ of TconNet are both set to 0.1. The 1-shot and 5-shot experiments are done with the same settings.

## 4. Study on Mixture Weight Vectors

### 4.1. Principal Component Analysis of Mixture Weight Vectors

We analyze the mixture weight vectors in the combined feature space constructed by query samples. The 5-shot 200+5 *tiered*ImageNet experiment is also considered. First, we did a principal component analysis (PCA) of 500 query samples for a test episode. The magnitude of the projected mixture weight vector to each principal component is computed in order to figure out how much power of mixture weight vector aligns with each principal component. In Fig. 1, it is shown that the magnitudes of mixture weight vectors for a backbone and MetaCNN are focused mainly on the first 150 or so principal axes of query samples. Consequently, the mixture weight vectors aim much of their attention at the critical dimensions where query samples show larger variances.

### 4.2. Magnitude Analysis

Fig. 2 shows the magnitude of the mixture weight vectors $\omega_{\text{pre}}$ and $\omega_{\text{meta}}$ during the meta-training phase. We consider the 5-shot 200+5 *tiered*ImageNet case. The mixture weight vector for the feature from the pretrained backbone is seen to have a larger magnitude than the weight vector for the feature from the MetaCNN. The magnitude of $\omega_{\text{pre}}$ fluctuates and slightly decreases until the learning rate decay at 4,000 episodes. Afterwards, the magnitude gradually increases and eventually saturates. For the mixture weight vector $\omega_{\text{meta}}$ from the MetaCNN, the magnitude steadily increases over time and also saturates soon. An obvious observation is that the magnitudes of the mixture weight vectors become quite stable before the meta-training phase is over. Also, given the difference in the magnitudes between the two weight vectors, we conjecture that XtarNet more actively utilizes the feature from the pretrained backbone, while also significantly leveraging the feature from MetaCNN.

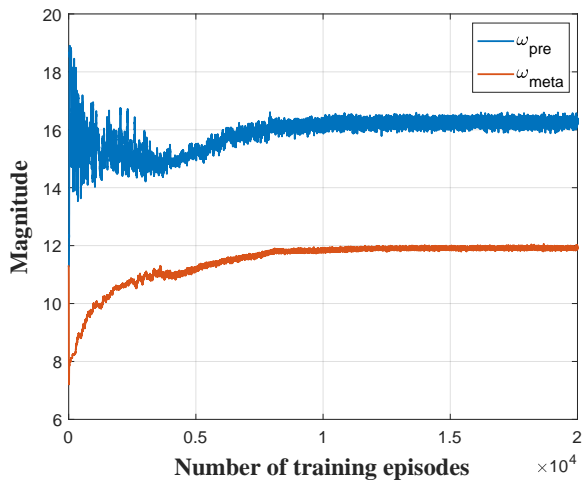

Figure 2. Magnitude of mixture weight vectors during the meta-training phase

Table 1. Dataset statistics of *mini*ImageNet

| Classes | Purpose | Split | $N_{\text{classes}}$ | $N_{\text{samples}}$ |
|---|---|---|---|---|
| Base ($\mathcal{D}_{\text{base}}$) | Train | Train-Train ($\mathcal{D}_{\text{base/train}}$) | 64 | 38,400 |
| | Val | Train-Val ($\mathcal{D}_{\text{base/val}}$) | 64 | 18,748 |
| | Test | Train-Test ($\mathcal{D}_{\text{base/test}}$) | 64 | 19,200 |
| Novel ($\mathcal{D}_{\text{novel}}$) | Train | Train-Train ($\mathcal{D}_{\text{novel/train}} \leftarrow \mathcal{D}_{\text{base/train}}$) | 64 | 38,400 |
| | Val | Val ($\mathcal{D}_{\text{novel/val}}$) | 16 | 9,600 |
| | Test | Test ($\mathcal{D}_{\text{novel/test}}$) | 20 | 12,000 |

Table 2. Dataset statistics of *tiered*ImageNet

| | Purpose | Split | $N_{\text{classes}}$ | $N_{\text{samples}}$ |
|---|---|---|---|---|
| Base ($\mathcal{D}_{\text{base}}$) | Train | Train-A-Train ($\mathcal{D}_{\text{base/train}}$) | 200 | 203,751 |
| | Val | Train-A-Val ($\mathcal{D}_{\text{base/val}}$) | 200 | 25,460 |
| | Test | Train-A-Test ($\mathcal{D}_{\text{base/test}}$) | 200 | 25,488 |
| Novel ($\mathcal{D}_{\text{novel}}$) | Train | Train-B ($\mathcal{D}_{\text{novel/train}}$) | 151 | 193,996 |
| | Val | Val ($\mathcal{D}_{\text{novel/val}}$) | 97 | 124,261 |
| | Test | Test ($\mathcal{D}_{\text{novel/test}}$) | 160 | 206,209 |

## 5. Dataset Statistics

In this section, we present the detailed statistics of the *mini*ImageNet and *tiered*ImageNet datasets that we use in the incremental few-shot learning tests. We follow the data splits used in prior works of (Gidaris & Komodakis, 2018) and (Ren et al., 2019). The data splits of *mini*ImageNet are presented in Table 1. For the *mini*ImageNet experiments, the training set $\mathcal{D}_{\text{base/train}}$ (denoted by 'Train-Train'), which is for pretraining of base classes, is reused as the training set $\mathcal{D}_{\text{novel/train}}$ (denoted by 'Train-Val') for novel classes. Since the data split $\mathcal{D}_{\text{novel/train}}$ contains only base classes, some randomly selected base classes are used as fake novel classes. This process is suggested in the prior work (Gidaris & Komodakis, 2018) and (Ren et al., 2019).

On the other hand, for the *tiered*ImageNet experiments, the $\mathcal{D}_{\text{base/train}}$ and $\mathcal{D}_{\text{novel/train}}$ are different. The original training split of *tiered*ImageNet which contains 351 classes is divided into two splits of 'Train-A' and 'Train-B' containing disjoint classes. The 'Train-A' is then further divided into training/validation/test sets of 'Train-A-Train', 'Train-A-Val' and 'Train-A-Test' with disjoint samples. These splits are for base classes. For novel categories, 151 extra classes of a set 'Train-B' are used for novel classes in the meta-training phase. For validation and test, sets denoted by 'Val' and 'Test' are used respectively. These splits for novel categories contain disjoint classes.

## 6. Few-shot Classification Accuracies

Mainly out of curiosity, we also evaluate few-shot classification accuracies of XtarNet, which is trained as an incremental few-shot learner supposed to handle both base and novel categories. The accuracy rates on *mini*ImageNet datasets, as shown in Table 3, are all from previously published results, except for our XtarNet.

It can be seen that the known incremental few-shot learning methods, LwoF and Attention Attractor Networks, show significantly degraded performance compared to the ResNet12-based learners specialized to few-shot learning. Compared to these prior incremental few-shot learners, our XtarNet shows better accuracies, indicating relatively strong few-shot learning performance although its design aimed at incremental few-shot learning tasks.

## 7. Algorithmic Description

Algorithm 1 shows a pseudocode-style description of the meta-training stage for XtarNet. For setting the initial novel classifier weights, we adopt meta-learnable references of TapNet (Yoon et al., 2019). Also, the classification of queries is carried out in the task-adaptive projection space.

## 8. Time Complexity

In Tables 1 and 2 of the main paper, XtarNet shows state-of-the-art accuracies on incremental few-shot learning tasks. AANs show the second best performance. We further remark that our XtarNet has significant advantage in time

Table 3. Few-shot classification accuracies for 5-way *mini*ImageNet

| Methods | backbone | 5-way *mini*ImageNet 1-shot | 5-shot |
|---|---|---|---|
| **MAML** (Finn et al., 2017) | Conv4 | $48.70 \pm 1.84\%$ | $63.15 \pm 0.91\%$ |
| **Prototypical Nets** (Snell et al., 2017) | Conv4 | $49.42 \pm 0.78\%$ | $68.20 \pm 0.66\%$ |
| **Transductive Propagation Nets** (Liu et al., 2018) | Conv4 | $55.51 \pm 0.86\%$ | $69.86 \pm 0.65\%$ |
| **TADAM** (Oreshkin et al., 2018) | ResNet12 | $58.5 \pm 0.3\%$ | $76.7 \pm 0.3\%$ |
| **TapNet** (Yoon et al., 2019) | ResNet12 | $61.65 \pm 0.15\%$ | $76.36 \pm 0.10\%$ |
| **MetaOpt** (Lee et al., 2019) | ResNet12 | $62.64 \pm 0.62\%$ | $78.63 \pm 0.46\%$ |
| **Cross Attention Nets*** (Hou et al., 2019) | ResNet12 | $63.85 \pm 0.48\%$ | $79.44 \pm 0.34\%$ |
| **LwoF** (Gidaris & Komodakis, 2018) | ResNet12 | $55.45 \pm 0.89\%$ | $70.92 \pm 0.35\%$ |
| **Attention Attractor Networks** (Ren et al., 2019) | ResNet12 | $55.75 \pm 0.51\%$ | $70.14 \pm 0.44\%$ |
| **XtarNet** (Ours) | ResNet12 | $57.29 \pm 0.54\%$ | $75.39 \pm 0.43\%$ |

*The Cross Attention Networks version with high-confidence query samples yields higher accuracy but is not shown here for fairness reasons.

Table 4. Results of combining AAN with our TAR for *tiered*ImageNet

| Methods | 1-shot Accuracy | $\Delta$ | 5-shot Accuracy | $\Delta$ |
|---|---|---|---|---|
| **Attention Attractor Networks*** (Ren et al., 2019) | $56.11 \pm 0.33\%$ | $-6.11\%$ | $65.52 \pm 0.31\%$ | $-4.48\%$ |
| **Proposed method with AAN** | $60.23 \pm 0.36\%$ | $-2.03\%$ | $68.06 \pm 0.33\%$ | $-1.72\%$ |

*The Attention Attractor Networks results reflect the reported numbers of (Ren et al., 2019).

complexity over AANs. The meta-trained modules in XtarNet operate without any recursive optimization procedures, whereas AANs require several tens of inner-loop optimization steps for obtaining novel classifier weights. See Step 6 through 9 in Algorithm 1 of (Ren et al., 2019). This recursive process in AANs causes a substantial latency relative to XtarNet during query processing in both training and inference. Furthermore, during training, AANs actually employ additional inner-loop for computing gradients based on the recurrent back-propagation (RBP) algorithm. Steps 15 through 17 of Algorithm 1 of (Ren et al., 2019) represent the second inner loop. This further slows down meta-training of the learner relative to our method.

## 9. Additional Experiments for Combining Attention Attractor Network with TAR

In an effort to further evaluate the expandability of the proposed TAR, we run our method on top of AANs (Ren et al., 2019) for *tiered*ImageNet as shown in Table 4. As mentioned in the main paper, combining AANs with our proposed method is possible, although the combining strategy is not straightforward. The most challenging part is optimizing all of our modules along with the meta-learnable parameters of AANs integrated with the inner loop optimization procedure. In AANs, the attention-based regularizer is meta-trained to regularize a novel classifier in a way that novel classes predictions do not interfere the base classes during the inner loop optimization for each episode. Our strategy is to use TAR constructed by MetaCNN and MergeNet, and the base classifer conditioned by TconNet for computing this regularization term. We drop the meta-module for conditioning novel classifier $h_\lambda$ for simplicity. Naively computing the gradients with respect to all of these modules during the RBP step does not work well in practice, perhaps due to the large number of parameters to be optimized. In order to get around this computational burden, we compute the gradients only with respect to the meta parameters of AANs in the RBP step, and then backpropagate with respect to our meta-modules outside of the inner loop independently, given a novel classifier prepared from AANs. We ran experiment on the same sized backbone architecture used in (Ren et al., 2019) to gauge the effectiveness of our method. As shown in Table 4, we obtained a $68.06\%$ in accuracy and $-1.72\%$ in $\Delta$ value for the 5-shot tieredImageNet experiments. For the 1-shot case, TAR combined with AANs shows $60.23\%$ in accuracy and $-2.03\%$ in $\Delta$ value. These are considerable gains in both accuracies and $\Delta$ values compared to AANs alone.

**Algorithm 1** Description of episodic learning during meta-training of XtarNet. The pretrained backbone and base classification weights are prepared before meta-training begins.

**Input**: Backbone $f_\theta$; base classifier weights $\{\mathbf{w}_i\}_{i=1}^{N_b}$; MetaCNN $g$; MergeNet $r_{\text{pre}}$ and $r_{\text{meta}}$; TconNet $h_\gamma$, $h_\beta$ and $h_\lambda$; novel classifier weights $\{\mathbf{w}_i\}_{i=N_b+1}^{N_b+N}$

1: **for** $t = 1 \cdots T$ **do**
2:    $L_{\text{train}} \leftarrow 0$
3:    $\mathcal{S}, \mathcal{Q} \leftarrow \text{GetEpisode}(\mathcal{D}_{\text{base/train}}, \mathcal{D}_{\text{novel/train}})$     // support set $\mathcal{S}$ and query set $\mathcal{Q}$
4:    $\mathbf{c} \leftarrow \frac{1}{|\mathcal{S}|} \sum_{\mathbf{x} \in \mathcal{S}} [f_\theta(\mathbf{x}), g(a_\theta(\mathbf{x}))]$     // $a_\theta(\mathbf{x})$ is an intermediate layer output of $f_\theta(\mathbf{x})$.
5:    $\boldsymbol{\omega}_{\text{pre}} \leftarrow r_{\text{pre}}(\mathbf{c})$, $\boldsymbol{\omega}_{\text{meta}} \leftarrow r_{\text{meta}}(\mathbf{c})$
6:    **for** $k = N_b + 1, \cdots, N_b + N$ **do**
7:       $\mathbf{c}_k^* \leftarrow \frac{1}{|\mathcal{S}_k|} \sum_{\mathbf{x} \in \mathcal{S}_k} [\boldsymbol{\omega}_{\text{pre}} \odot f_\theta(\mathbf{x}) + \boldsymbol{\omega}_{\text{meta}} \odot g(a_\theta(\mathbf{x}))]$     // per-class averages of combined features
8:    **end for**
9:    $\mathbf{c}^* \leftarrow \frac{1}{N} \sum_{k=N_b+1}^{N_b+N} \mathbf{c}_k^*$
10:   **for** $i = 1, \cdots, N_b$ **do**
11:      $\mathbf{w}_i^* \leftarrow (\mathbb{1} + a h_\gamma(\mathbf{c}^*)) \odot \mathbf{w}_i + b h_\beta(\mathbf{c}^*)$     // hyperparameters $a$ and $b$ are set to 0.1.
12:   **end for**
13:   **for** $k = N_b + 1 \cdots N_b + N$ **do**
14:      **for** $i = 1, \cdots, N_b$ **do**
15:         $\sigma_k^i \leftarrow \mathbf{c}_k^* \cdot \mathbf{w}_i^*$
16:      **end for**
17:      $\boldsymbol{\sigma}_k \leftarrow [\sigma_k^1, \cdots, \sigma_k^{N_b}]$
18:      $\boldsymbol{\lambda}_k \leftarrow h_\lambda(\boldsymbol{\sigma}_k)$
19:   **end for**
20:   **for** $i = N_b + 1 \cdots N_b + N$ **do**
21:      $\mathbf{w}_i^* = \mathbf{w}_i - \sum_{j=1}^{N_b} \frac{\exp(\lambda_i^j)}{\sum_l \exp(\lambda_i^l)} \mathbf{w}_j^*$     // $\lambda_i^j$ is $i^{\text{th}}$ element of $\boldsymbol{\lambda}_k$.
22:   **end for**
23:   **for** $k = N_b + 1, \cdots, N_b + N$ **do**
24:      $\boldsymbol{\epsilon}_i \leftarrow \mathbf{w}_k^*/||\mathbf{w}_k^*|| - \mathbf{c}_k^*/||\mathbf{c}_k^*||$
25:   **end for**
26:   $\mathbf{M} \leftarrow \text{null}\{[\boldsymbol{\epsilon}_{N_b+1}, \cdots, \boldsymbol{\epsilon}_{N_b+N}]\}$     // task-adaptive projection space
27:   **for** $i = 1, \cdots, N_b + N$ **do**
28:      **for** $(\mathbf{x}, y = i) \in \mathcal{Q}$ **do**
29:         $D_i(\mathbf{x}) = d\left(\mathbf{w}_i^* \mathbf{M}, [\boldsymbol{\omega}_{\text{pre}} \odot f_\theta(\mathbf{x}) + \boldsymbol{\omega}_{\text{meta}} \odot g(a_\theta(\mathbf{x}))]\mathbf{M}\right)$     // $d(\cdot, \cdot)$: Euclidean distance
30:         $L_{\text{train}} = L_{\text{train}} + \frac{1}{|\mathcal{Q}|}\left[D_i(\mathbf{x}) + \log \sum_l \exp\left(-D_l(\mathbf{x})\right)\right]$     // cross entropy loss
31:      **end for**
32:   **end for**
33:   Update all parameters of $g$, $r_{\text{pre}}$, $r_{\text{meta}}$, $h_\gamma$, $h_\beta$ $h_\lambda$, $\{\mathbf{w}_i\}_{i=N_b+1}^{N_b+N}$ minimizing $L_{\text{train}}$ via optimizer
34: **end for**



\end{document}